\documentclass[a4paper,conference]{IEEEtran}
\IEEEoverridecommandlockouts
\usepackage{cite}
\usepackage{amsmath,amssymb,amsfonts}
\usepackage{algorithmic}
\usepackage{graphicx}
\usepackage{makecell}
\usepackage{multirow}
\usepackage{textcomp}
\usepackage{orcidlink}

\usepackage{hyperref}
\hypersetup{
    colorlinks=true,
    linkcolor={blue!100!black},
    citecolor={blue!100!black},
    urlcolor={blue!80!black},
    pdftitle={Low-Cost Machine Vision System for Sorting Green Lentils (Lens Culinaris) Based on Pneumatic Ejection and Deep Learning},
}

\def\BibTeX{{\rm B\kern-.05em{\sc i\kern-.025em b}\kern-.08em
    T\kern-.1667em\lower.7ex\hbox{E}\kern-.125emX}}

\begin{document}

\title{Low-Cost Machine Vision System for Sorting Green Lentils (\textit{Lens Culinaris}) Based on\\ Pneumatic Ejection and Deep Learning}

\author{\IEEEauthorblockN{Davy Rojas Yana \orcidlink{0009-0003-8520-9929} and Edwin Salcedo \orcidlink{0000-0001-8970-8838}}
\IEEEauthorblockA{\textit{Department of Mechatronics Engineering}\\ \textit{Universidad Católica Boliviana ``San Pablo'', La Paz, Bolivia} \\
{\tt\small \{davy.rojasy,ingedwinsalcedo\}@gmail.com}}}

\maketitle

\begin{abstract}

This paper presents the design, development, and evaluation of a dynamic grain classification system for green lentils (\textit{Lens Culinaris}), which leverages computer vision and pneumatic ejection. The system integrates a YOLOv8-based detection model that identifies and locates grains on a conveyor belt, together with a second YOLOv8-based classification model that categorises grains into six classes: Good, Yellow, Broken, Peeled, Dotted, and Reject. This two-stage YOLOv8 pipeline enables accurate, real-time, multi-class categorisation of lentils, implemented on a low-cost, modular hardware platform. The pneumatic ejection mechanism separates defective grains, while an Arduino-based control system coordinates real-time interaction between the vision system and mechanical components. The system operates effectively at a conveyor speed of 59~mm/s, achieving a grain separation accuracy of 87.2\%. Despite a limited processing rate of 8 grams per minute, the prototype demonstrates the potential of machine vision for grain sorting and provides a modular foundation for future enhancements. 

\end{abstract}

\begin{IEEEkeywords}
Green Lentil Classification, Lentil Quality Inspection, Machine Vision, Pneumatic Systems, Real-Time Sorting
\end{IEEEkeywords}

\section{Introduction}

Lentil (\textit{Lens culinaris Medik.}) is one of the most widely cultivated legumes globally, with Canada, India, Turkey and the United States being the main producing regions \cite{fao2025}. According to recent agricultural market data, global lentil production reached approximately 8.6 million metric tonnes in 2023 \cite{indexbox2025}. Given the substantial supply of this grain, its appearance—particularly colour and shape—is of greatest interest to consumers and is, therefore, one of the main factors determining its value \cite{shahin2003}. Appearance can be assessed as the combined effect of colour, colour uniformity, discolouration, and shape defects.

Green lentils are the primary variety consumed in the region, exhibiting the range of visual characteristics as previously described. This variability makes accurate classification based on appearance particularly challenging. However, precise identification is essential, as the inability to distinguish between different grain types can diminish the product's market value. Implementing appearance-based grain classification ensures that only high-quality lentils reach consumers—an outcome that is vital for meeting market expectations, competing with premium imported products, and potentially enabling export opportunities supported by consistent quality.

At present, lentil classification in our context is typically carried out manually. This process, based on visual assessment of physical appearance, is both costly and inefficient, as a person can evaluate only 10 to 50 grams of grains per minute \cite{wan2003}. Moreover, human evaluation is often inconsistent and subjective \cite{shahin2003}, and performing such repetitive tasks routinely can lead to significant physical and mental fatigue. In contrast, machine vision systems offer a more effective alternative, enabling rapid and accurate classification of grains \cite{salcedo2024}. These systems also help reduce the burden of repetitive labour, particularly for vulnerable groups of workers.

Unlike prior work on lentil quality classification, which has been limited to static or semi-automated inspection systems, this prototype integrates real-time deep learning (DL) inference with precision pneumatic actuation on a moving conveyor, offering a replicable, low-cost solution suitable for local agricultural contexts. Furthermore, the system is controlled by a YOLOv8-based artificial intelligence pipeline for real-time detection and sorting. This pipeline implements a multi-class classification strategy using an anchor-free YOLOv8 model trained on six detailed grain categories. This level of granularity in classification, combined with real-time deployment in a dynamic conveyor system, represents a methodological contribution to the field of agricultural machine vision.

\begin{figure*}[htbp]
  \centering
  \includegraphics[width=0.9\textwidth]{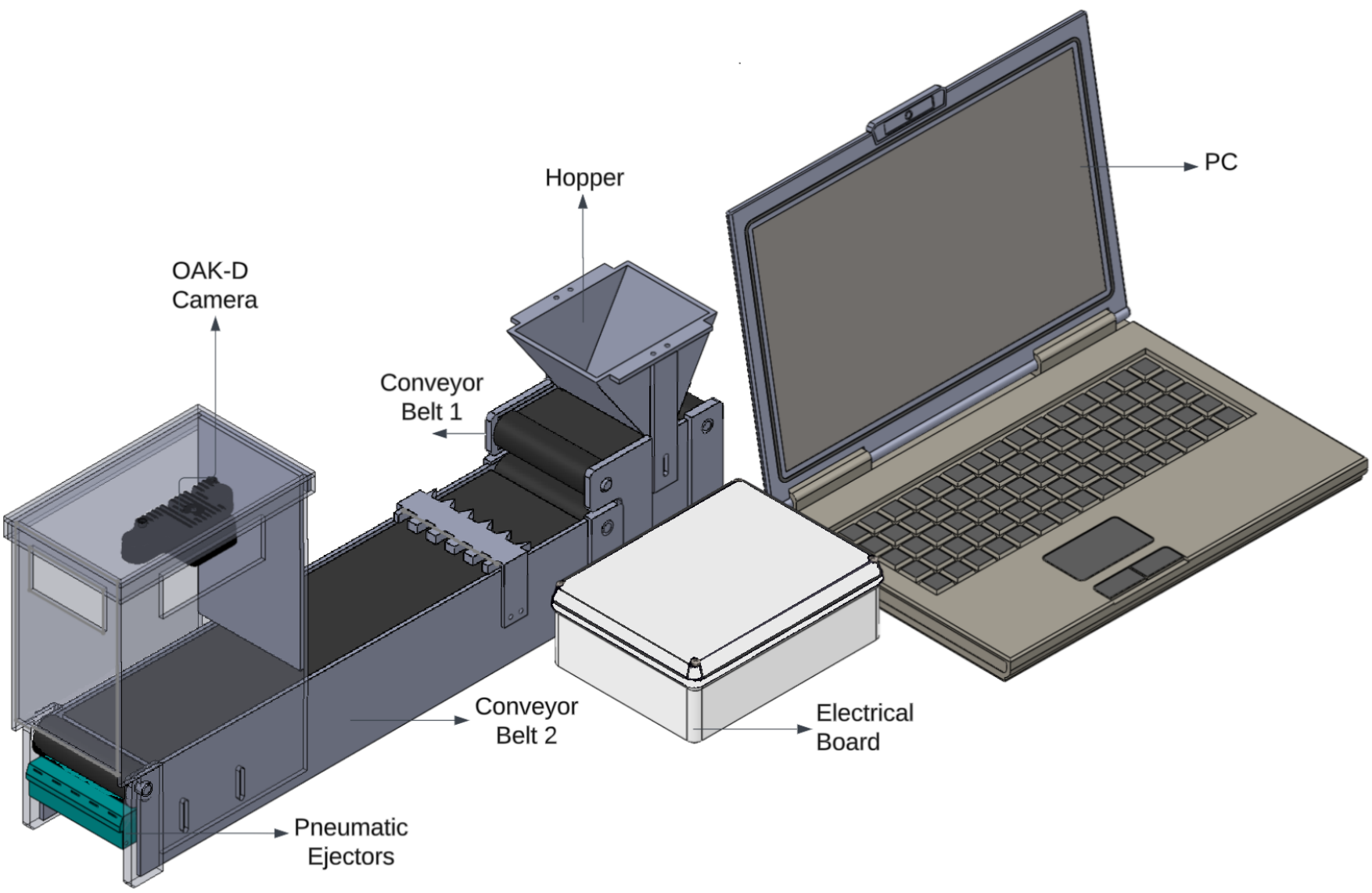}
  \caption{System design using the 3D CAD software SolidWorks.}
  \label{fig:prototype}
\end{figure*}

\section{Related Work}
In the development of grain grading systems, both academic and commercial solutions have addressed challenges similar to those presented in this paper. These solutions are distinguished by their focus on utilising computer vision, machine learning, and artificial intelligence technologies to enhance grain grading accuracy. Several notable works related to grain grading are highlighted below.

Chen et al. from South China Agricultural University \cite{chen2019} proposed a computer vision system coupled with Support Vector Machine (SVM) classifiers to identify defects in rice grains. Their methodology involves image segmentation to detect chalky, broken, or stained areas using distance constraints and edge detection algorithms. Furthermore, Del Coco et al. \cite{delcoco2022} developed a system that uses RGB images captured under controlled lighting conditions to assess the size and morphology of lentil grains. However, their approach is static and does not account for the dynamic conditions typical in industrial settings.

T. Pearson \cite{pearson2010} introduced a system based on complementary metal–oxide–semiconductor (CMOS) sensors and real-time processing with field-programmable gate array (FPGA) technology. This prototype enables the classification of grains during free fall, achieving relatively high throughput due to parallel processing. Recent trends have shifted towards DL-based classifiers, exemplified by a system developed at ACE University of Engineering in India \cite{kundu2022}, which uses a YOLOv5 model for seed classification. The system classifies seeds as either good or bad, utilising an expanded dataset created through image augmentation techniques. While this approach excels in autonomous learning, it focuses on a limited number of categories and operates under controlled conditions.

In the development of mechanical systems for grain ejection, Persak et al. \cite{persak2020} designed a system to classify transparent polycarbonate particles using a conveyor belt integrated with a camera. Defect separation was achieved through pneumatic nozzles controlled by classification algorithms based on k-Nearest Neighbors (k-NN). This system is particularly relevant for its integration of computer vision with pneumatic ejection, a concept directly applicable to dynamic grain classification.

\section{Materials and Methods}
The proposed grain sorting system integrates three key subsystems: mechanical, machine vision, and electronic. Each subsystem has been meticulously designed to operate in real time, enabling the detection, sorting, and separation of grains as they move through the system. Figure \ref{fig:prototype} shows the 3D design of the system prototype.

\subsection{Mechanical System}

The mechanical system comprises a conveyor belt for grain movement, a grain feeder belt, and an ejection system. The conveyor belt features a grain positioning width of 100 mm, divided into five rows spaced 20 mm apart, ensuring a uniform flow of grains. The total length of the conveyor belt is 400 mm, providing stabilization for grains as they fall from the feeder belt. The prototype is constructed using medium-density fibreboard (MDF) wood and polylactic acid (PLA) for 3D-printed components. A black rubber band is incorporated to create a contrast with the grains, facilitating  visual processing. The belt is powered by a NEMA17 stepper motor, which delivers the required torque and enables precise speed control.

\begin{figure}[htbp]
  \centering
  \includegraphics[width=.3\textwidth]{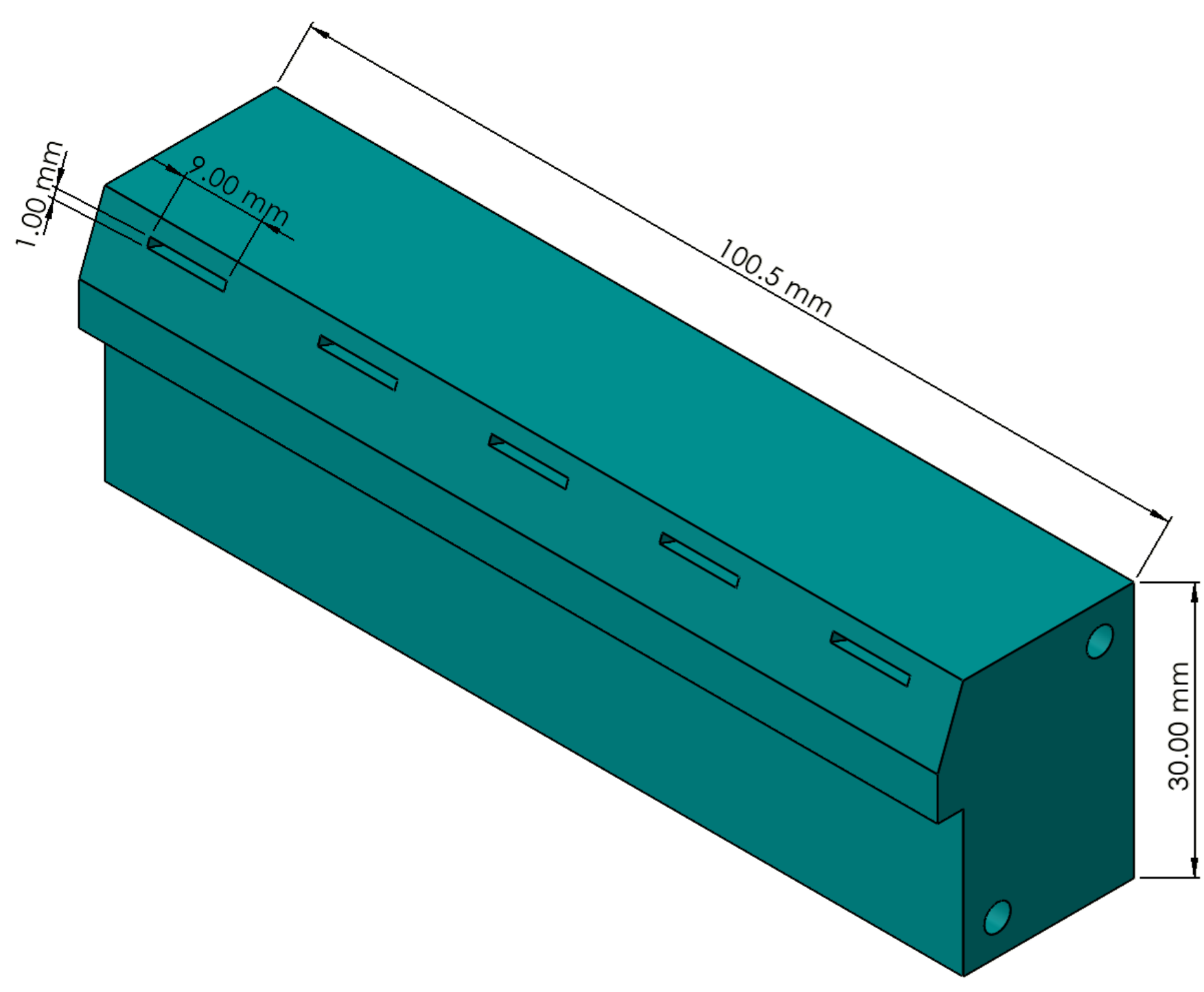}
  \caption{Nozzles employed in the pneumatic ejectors for grain sorting.}
  \label{fig:pneumatic}
\end{figure}

\begin{figure*}[htbp]
  \centering
  \includegraphics[width=\textwidth]{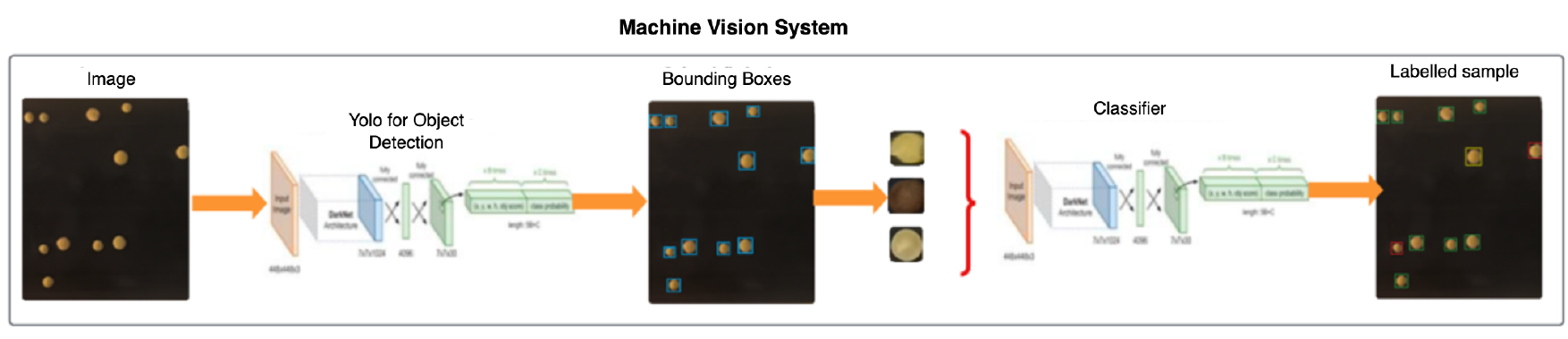}
  \caption{Pipeline of the machine vision system.}
  \label{fig:mv-system}
\end{figure*}

The feeding system features a hopper designed to hold up to 300 g of lentils, ensuring consistent flow to the conveyor belt. Positioned beneath the hopper, an additional feed conveyor belt evenly distributes the grains onto the main conveyor belt, with both conveyors synchronised by a separate stepper motor.

The ejection system is composed of pneumatic ejectors: five nozzles aligned at the end of the conveyor belt, responsible for separating defective grains. Each nozzle is connected to a solenoid valve within Matrix 890 Series valve assemblies (model OX899900C2KK), which are high-frequency valves capable of switching in less than 1 ms. The nozzles feature a rectangular cross-section of 1 x 9 mm and are angled at 20° relative to the vertical axis, optimising their alignment with the grains' fall and dynamic behaviour, as illustrated in Figure \ref{fig:pneumatic}.

\begin{figure}[htbp]
  \centering
  \includegraphics[width=.45\textwidth]{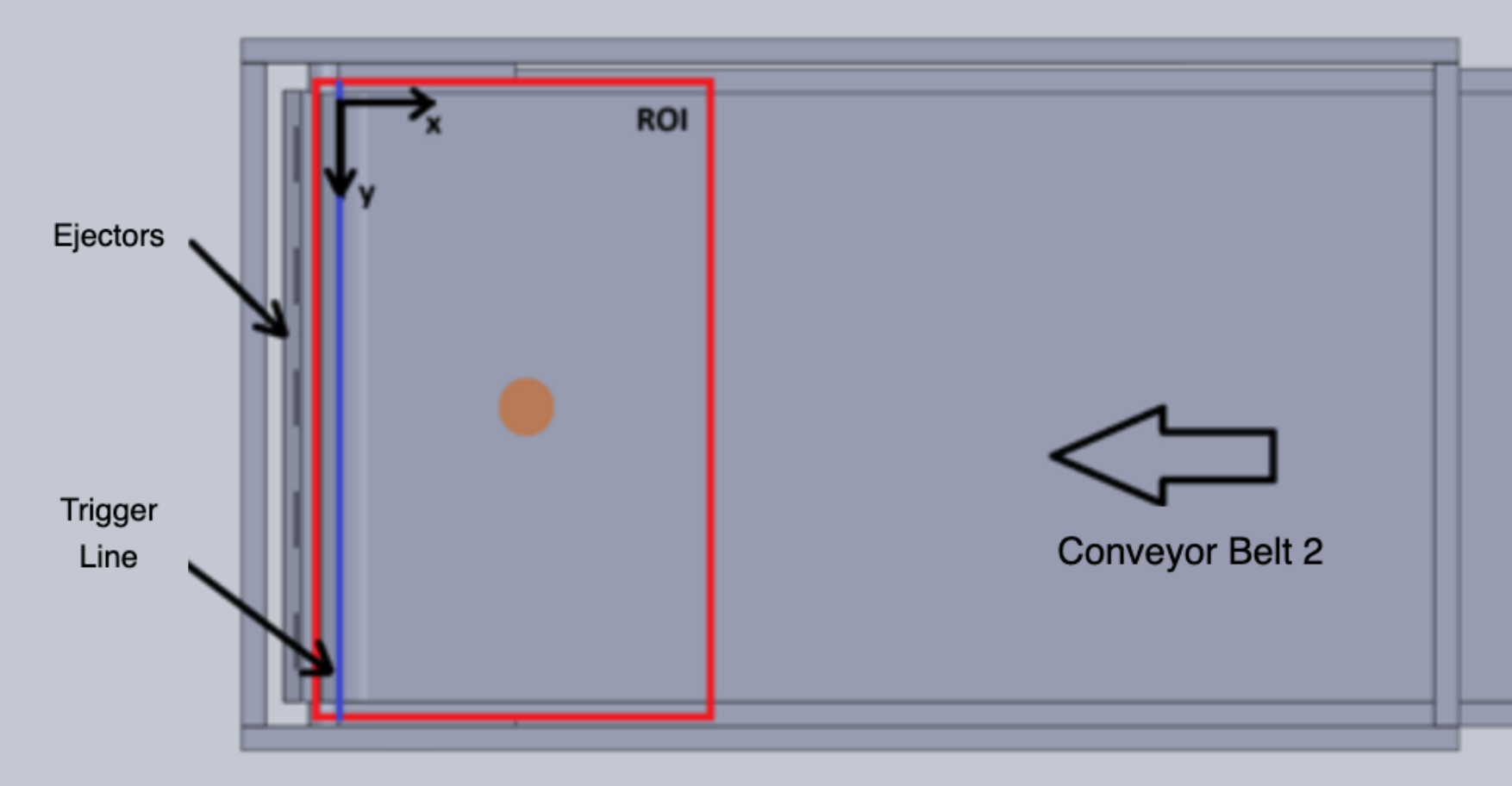}
  \caption{The process of determining which ejector is activated and when the grain is ejected.}
  \label{fig:ejection}
\end{figure}

\subsection{Machine Vision System}

This system, as shown in Figure~\ref{fig:mv-system}, employs a Luxonis OAK-D camera, with a resolution of 1920~$\times$~1080 pixels and a capture rate of 40 FPS, to capture images. Positioned 96 mm above the moving conveyor belt, the camera provides a vertical field of view of 100 mm. While we initially attempted to implement a single DL model for grain sorting, the best-performing approach was to use two YOLOv8 models, described as follows.

First, a YOLOv8s model (hereafter referred to as ``YOLOv8 for Detection'') detects grains, generating bounding boxes around each lentil grain. These boxes contain information on the position and size of the detected grains in the image. For classification, each cropped bounding box is resized to 32~$\times$~32 pixels and categorised by the ``YOLOv8 for Classification'' model (specifically YOLOv8m-cls) into one of six classes: Good, Yellow, Broken, Peeled, Dotted, or Refuse. Once classified, the data are processed to determine the row and the precise timing at which each grain will reach the ejection position. This timing triggers the activation of the corresponding ejector when the grain reaches the firing line in the image, as illustrated in Figure~\ref{fig:ejection}.

YOLOv8 (You Only Look Once version 8) is a state-of-the-art, anchor-free object detection model optimised for real-time inference. YOLOv8 was selected for this application due to its anchor-free design, which eliminates the need for predefined anchor boxes and simplifies the training process. Its efficient architecture and high detection accuracy make it particularly suitable for real-time tasks such as dynamic grain sorting. Additionally, its low computational demand enables deployment on embedded or edge-computing platforms \cite{terven2023}.

The architecture comprises three main components: a backbone for feature extraction, a neck for feature aggregation, and a head for final prediction. Given an input image $\mathbf{X} \in \mathbb{R}^{H \times W \times C}$, YOLOv8 predicts a set of bounding boxes $\{B_i\}$, objectness scores $\{s_i\}$, and class probabilities $\{p_{i,c}\}$, where $i$ indexes the detected objects and $c \in \{1, \dots, C\}$ denotes the class labels \cite{terven2023}. In this project, the backbone is based on EfficientNet, a convolutional neural network (CNN) architecture that efficiently extracts hierarchical representations from the input image:
\begin{equation}
\mathbf{F} = f_{\text{backbone}}(\mathbf{X}; \theta),
\end{equation}
where $\theta$ denotes the model parameters and $\mathbf{F}$ represents the extracted feature maps.

To enhance multi-scale representation, the feature maps are passed through a neck architecture that combines Feature Pyramid Networks (FPN) and Path Aggregation Networks (PANet):
\begin{equation}
\mathbf{F}_{\text{agg}} = f_{\text{neck}}(\mathbf{F}).
\end{equation}
This stage enriches the semantic information across different spatial resolutions, enabling the model to detect objects of varying sizes more effectively.

YOLOv8 employs an anchor-free detection head, which directly predicts the bounding box parameters—centre coordinates $(x_i, y_i)$, width $w_i$, and height $h_i$—along with the objectness score $s_i$ and class probabilities $\mathbf{p}_i$ coming from the activation function softmax:
\begin{equation}
\hat{B}_i = (x_i, y_i, w_i, h_i), \quad \hat{s}_i \in [0,1], \quad \hat{\mathbf{p}}_i = \text{softmax}(\mathbf{z}_i).
\end{equation}

The model is trained to minimise a composite loss function that balances localisation, objectness, and classification accuracy:
\begin{equation}
\mathcal{L} = \lambda_{\text{box}} \cdot \mathcal{L}_{\text{IoU}} + \lambda_{\text{obj}} \cdot \mathcal{L}_{\text{BCE}}^{\text{obj}} + \lambda_{\text{cls}} \cdot \mathcal{L}_{\text{BCE}}^{\text{cls}},
\end{equation}
where:
\begin{itemize}
  \item $\mathcal{L}_{\text{IoU}}$ is the bounding box regression loss (e.g., CIoU or DIoU),
  \item $\mathcal{L}_{\text{BCE}}^{\text{obj}}$ is the binary cross-entropy loss for objectness prediction,
  \item $\mathcal{L}_{\text{BCE}}^{\text{cls}}$ is the classification loss, often implemented as binary cross-entropy or focal loss,
  \item $\lambda_{\text{box}}, \lambda_{\text{obj}}, \lambda_{\text{cls}}$ are scalar weights that control the relative importance of each component.
\end{itemize}

During inference, YOLOv8 applies Non-Maximum Suppression (NMS) to remove redundant predictions. The Intersection over Union (IoU) between two boxes $B_i$ and $B_j$, also known as the Jaccard index \cite{salcedo2022}, is defined as:
\begin{equation}
\text{IoU}(B_i, B_j) = \frac{|B_i \cap B_j|}{|B_i \cup B_j|}.
\end{equation}
Detections with IoU above a predefined threshold (commonly 0.5) are suppressed in favour of those with higher confidence scores, ensuring one bounding box per object instance.

\subsection{Electronic System}

The electronic subsystem bridges the machine vision system with the mechanical actuators. The conveyor belts are controlled via an Arduino UNO development board, using Pololu A4988 drivers to regulate speed. The travel and feed belts have variable speeds, allowing for the adjustments during classification system tests. To control the pneumatic ejectors, an Arduino Nano board receives signals from the vision system and activates the corresponding valves through the Firmata protocol. Each ejector is triggered with precision based on the position and classification of the grain, ensuring accurate sorting and directing the grains to their appropriate path.

For central processing, a laptop equipped with an NVIDIA RTX 4070 GPU handles all computer vision tasks. Communication between the camera, Arduino boards, and the laptop is managed via USB 3.0, providing data transfer speeds. Additionally, the Firmata protocol is used for communication and control of the Arduino Nano microcontroller from the software, implemented in Python. This setup enables seamless, real-time interaction between the computer vision system and the mechanical actuators, such as the pneumatic ejectors.

The grains are fed into the system through the hopper and distributed onto the conveyor belt by the infeed conveyor. The camera captures images in real time and transmits them to the PC, where the YOLO Detection and YOLO Classification models process the images to identify and classify the grains accurately. The processed results are then used to determine the appropriate timing for activating the pneumatic ejectors. Defective grains are ejected, while those deemed acceptable continue along the conveyor to the collection area.

\section{Experimental Results}

The prototype developed for the dynamic grain classification system underwent experimental evaluation, providing valuable insights into the effectiveness of the machine vision pipeline and the overall system implementation.

\subsubsection{Object Detection}

\begin{figure}[htbp]
  \centering
  \includegraphics[width=.45\textwidth]{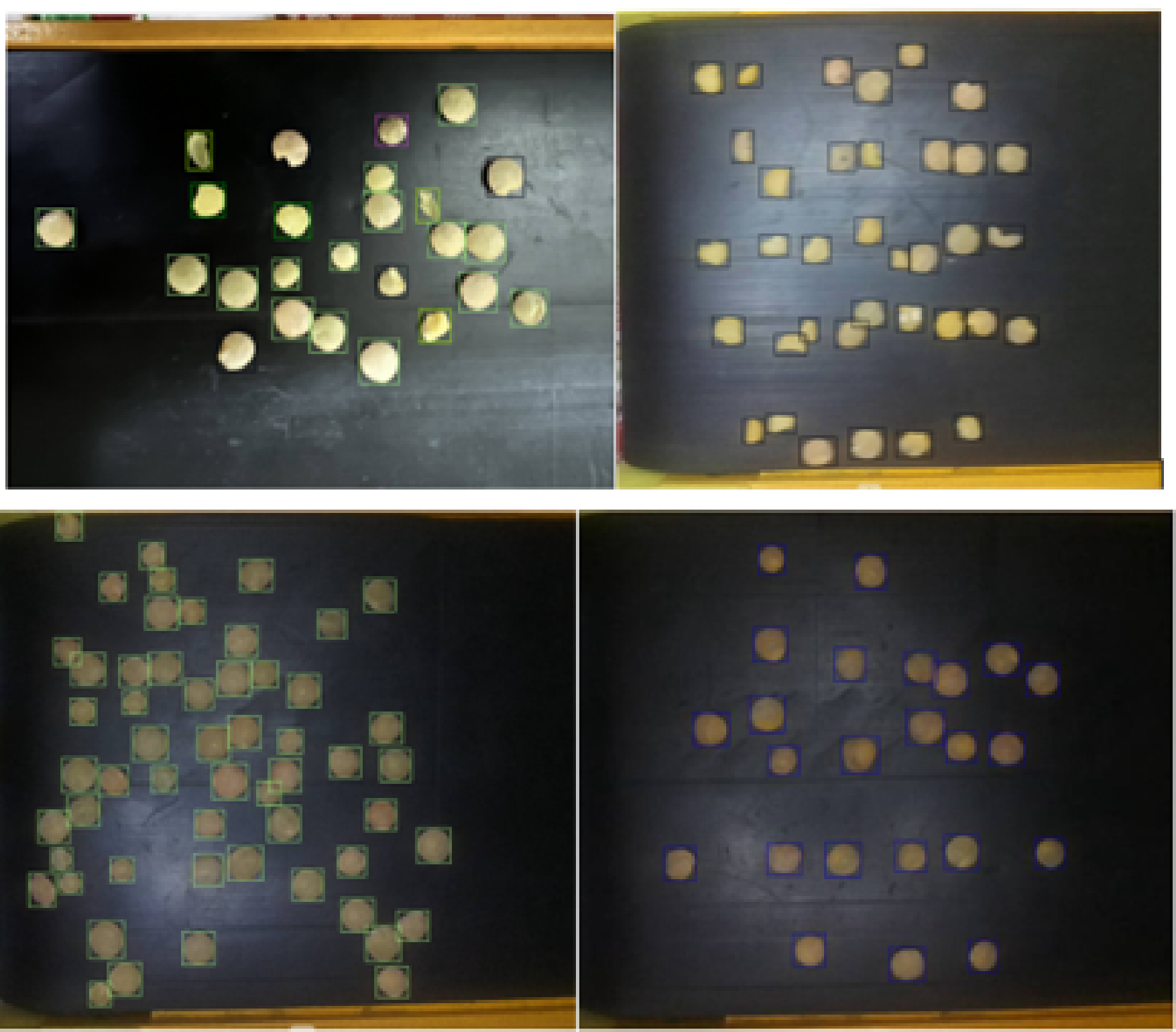}
  \caption{Annotated images to train the YOLOv8 for detection.}
  \label{fig:object-detection-annotated-samples}
\end{figure}

\begin{figure}[htbp]
  \centering
  \includegraphics[width=.45\textwidth]{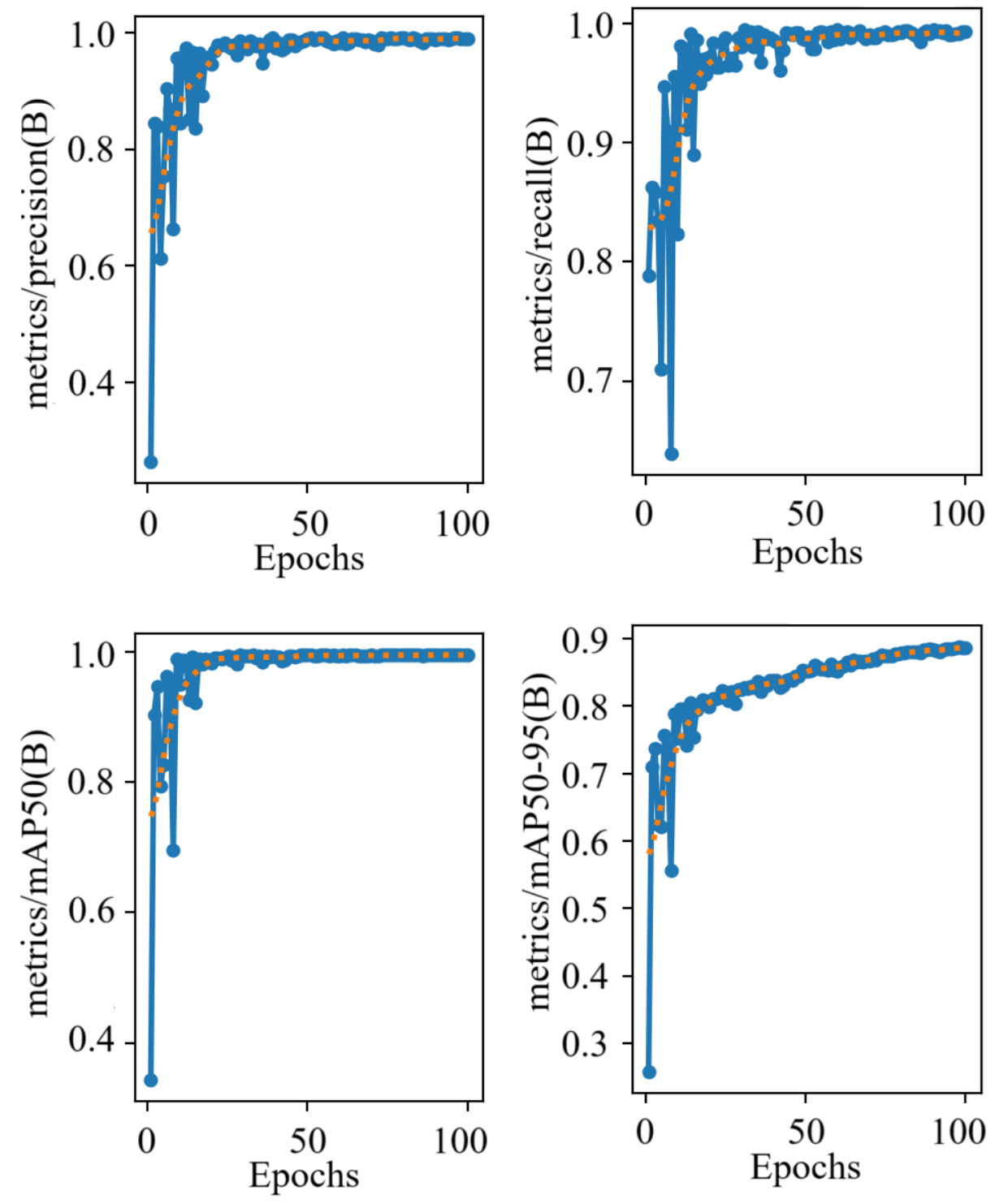}
  \caption{Metrics obtained while training the YOLOv8 for detection.}
  \label{fig:object-detection-metrics}
\end{figure}

We trained YOLOv8 for detection using the Lentil dataset, which consists of 382 photographs of lentil grains placed on the prototype’s conveyor belt while it was in motion. Sample images are shown in Figure~\ref{fig:object-detection-annotated-samples}. Each image was manually annotated using the CVAT tool~\cite{cvat2025}, with bounding boxes drawn around each grain and its corresponding class label specified. The dataset was divided into two subsets: 80\% for training and 20\% for validation, ensuring a robust evaluation of model performance while reducing the risk of overfitting. The following hyperparameters were used: batch size of 16, learning rate of 0.01, and 100 training epochs. The final model achieved an average mAP50–95 of 0.8872 and an mAP50 of 0.99466, demonstrating strong performance in grain detection. Figure~\ref{fig:object-detection-metrics} shows the YOLO detection model’s training curves for precision and loss across 100 epochs. The steady convergence and low final loss indicate effective learning and low overfitting risk.

\subsubsection{Lentil Quality Classification}

The initial grain classification results from the system deployment were suboptimal, prompting the decision to implement a dedicated YOLO model solely for grain classification. The YOLOv8 for  detection remained in use to locate grains within the image using bounding boxes, while the images within these boxes were sent to the YOLOv8 classification model to improve classification accuracy.

\begin{figure}[htbp]
  \centering
  \includegraphics[width=.45\textwidth]{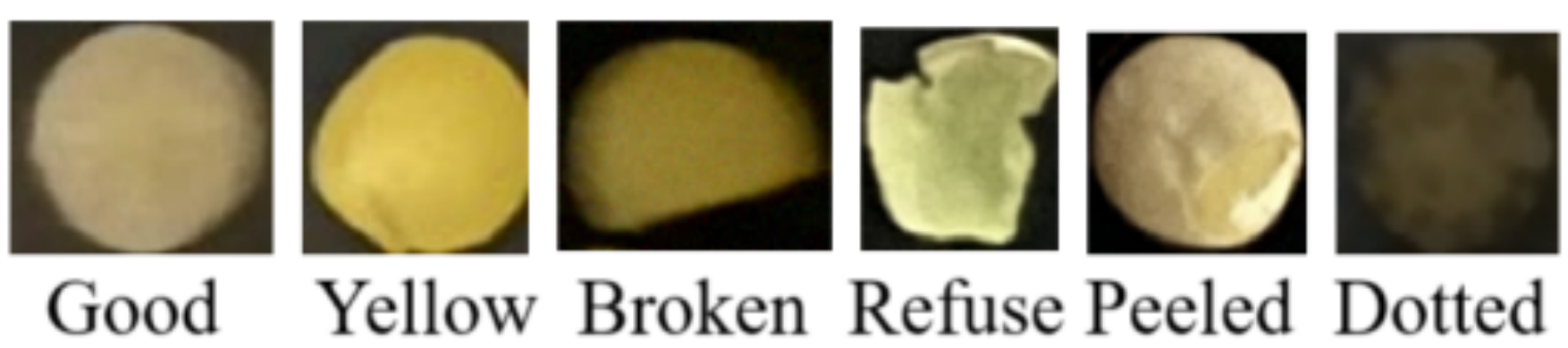}
  \caption{Sample images from the dataset for lentil classification.}
  \label{fig:category-samples}
\end{figure}

The Lentil dataset was used to create the training dataset for the classification model. From the detection model, images within the bounding boxes were cropped and segmented according to their respective classes. During segmentation, an imbalance in the dataset was identified. To address this, images were augmented through techniques such as resizing, rotation, and colour variation, ensuring a balanced dataset. As a result, a final dataset of 2,100 images per class was achieved. A sample of these images is shown in Figure \ref{fig:category-samples}.

For model training, the dataset was divided into two subsets: 80\% for training and 20\% for testing. To evaluate the model’s performance effectively, the following hyperparameters were used: Batch Size: 16, Learning Rate: 0.01, and Epochs: 50. After each epoch, performance metrics, including precision and loss, were recorded, as shown in Figure 8. The final model achieved a top-1 precision of 0.892 and a top-5 precision of 1.00, indicating strong performance in grain classification.

\subsection{System's Classification Performance}

To assess the performance of the integrated system, shown in Figure \ref{fig:demo}, a practical experiment was conducted to evaluate both the accuracy of the detection and classification models, as well as the effectiveness of the ejection system in segregating defective grains.

A controlled mixture of grains was prepared with the following distribution: 50 good grains (Good class), 10 yellow grains (Yellow class), 10 broken grains (Broken class), 10 foreign objects (Reject class), 10 peeled grains (Peeled class), and 10 dotted grains (Dotted class). The objective was to ensure that grains not classified as ``Good'' were separated by the pneumatic ejector. Ten experimental simulations were conducted, yielding an average classification accuracy of 87.2\%.

\begin{figure}[htbp]
  \centering
  \includegraphics[width=.45\textwidth]{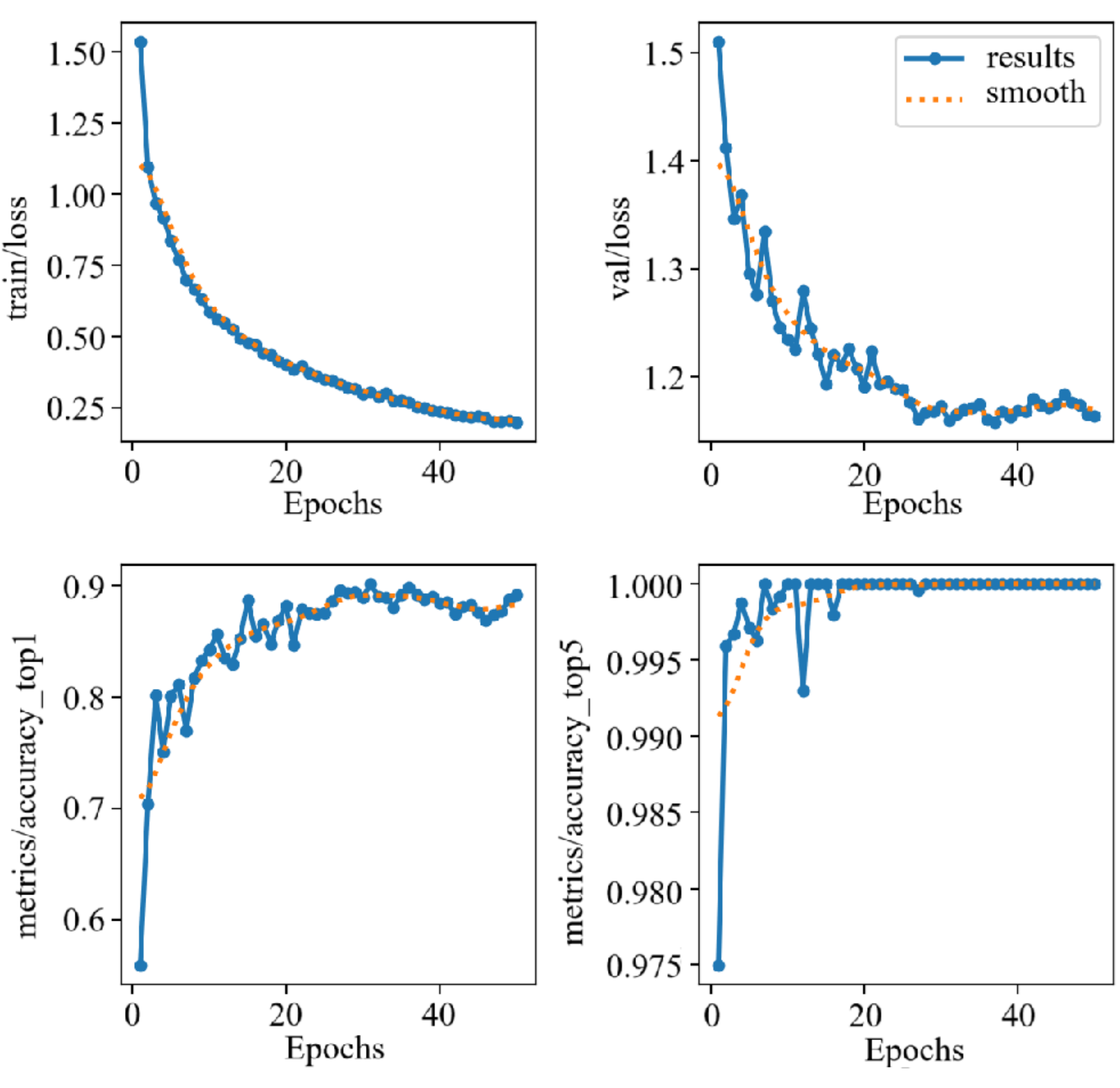}
  \caption{YOLO model training results for classification.}
  \label{fig:classification-metrics}
\end{figure}

\begin{figure}[htbp]
  \centering
  \includegraphics[width=.5\textwidth]{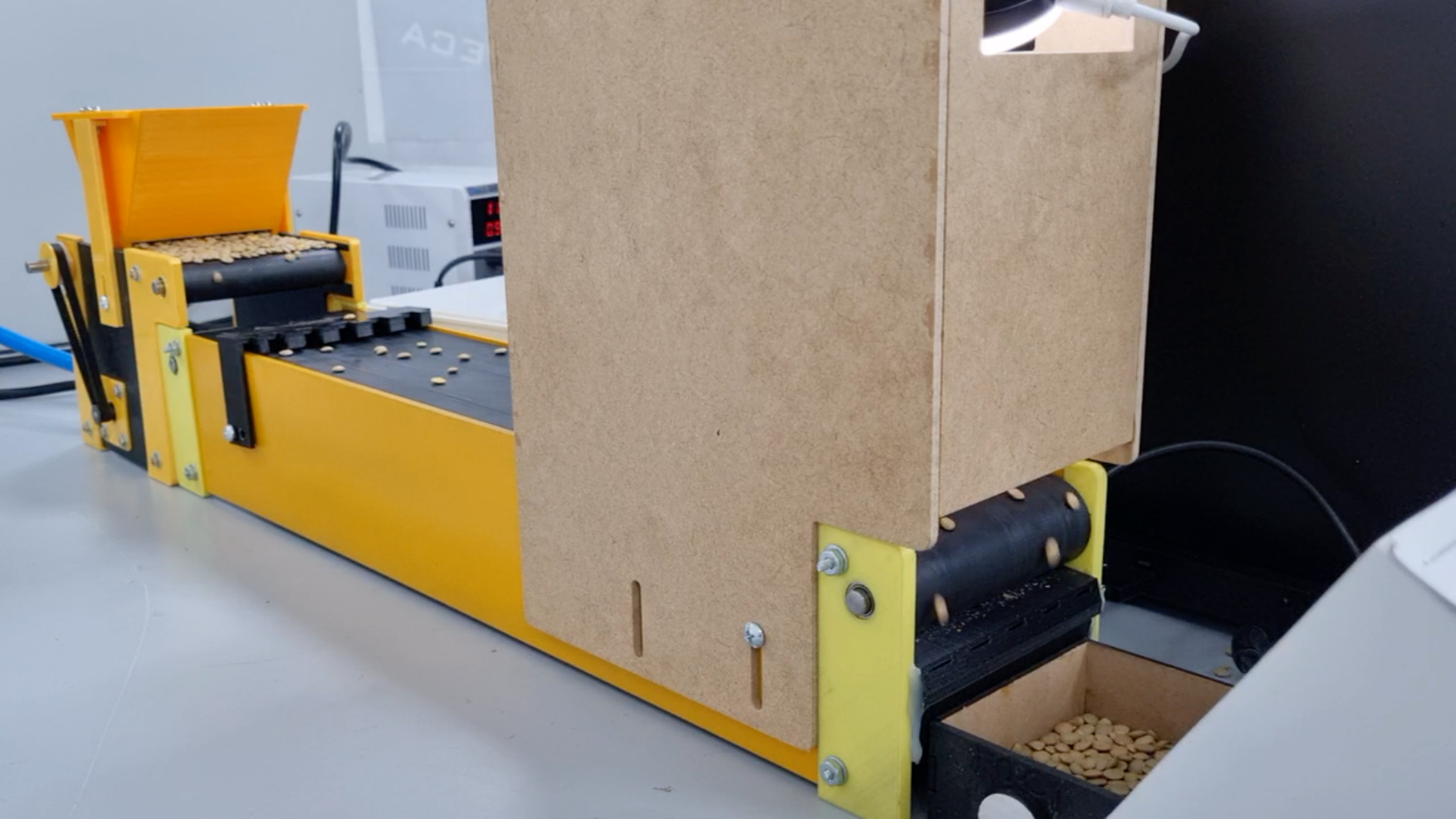}
  \caption{Final prototype. A demonstration video is included here \cite{authors2025}.}
  \label{fig:demo}
\end{figure}

\section{Discussion}

The proposed system achieved 87.2\% accuracy in separating good grains under controlled conditions, with a processing speed of approximately 8 grams per minute. While these results are acceptable for a prototype, there are several opportunities for improvement. Accuracy could be increased through the use of optimised models, such as quantised versions of YOLOv8, without significantly affecting real-time performance. The speed of the system is currently limited by the synchronisation between visual processing and pneumatic actuation, which could be improved through the use of higher-frequency controllers or embedded devices with accelerated inference capabilities. It is also suggested that the mechanical layout be reviewed: currently, the camera is oriented parallel to the horizontal plane, while the ejector nozzles are tilted 20° from the vertical axis, creating a misalignment between the trigger line and the ejection position. A viable alternative would be to reposition the camera so that it is aligned with the nozzles, thus eliminating the time lag between detection and ejection.

\begin{figure}[htbp]
  \centering
  \includegraphics[width=.48\textwidth]{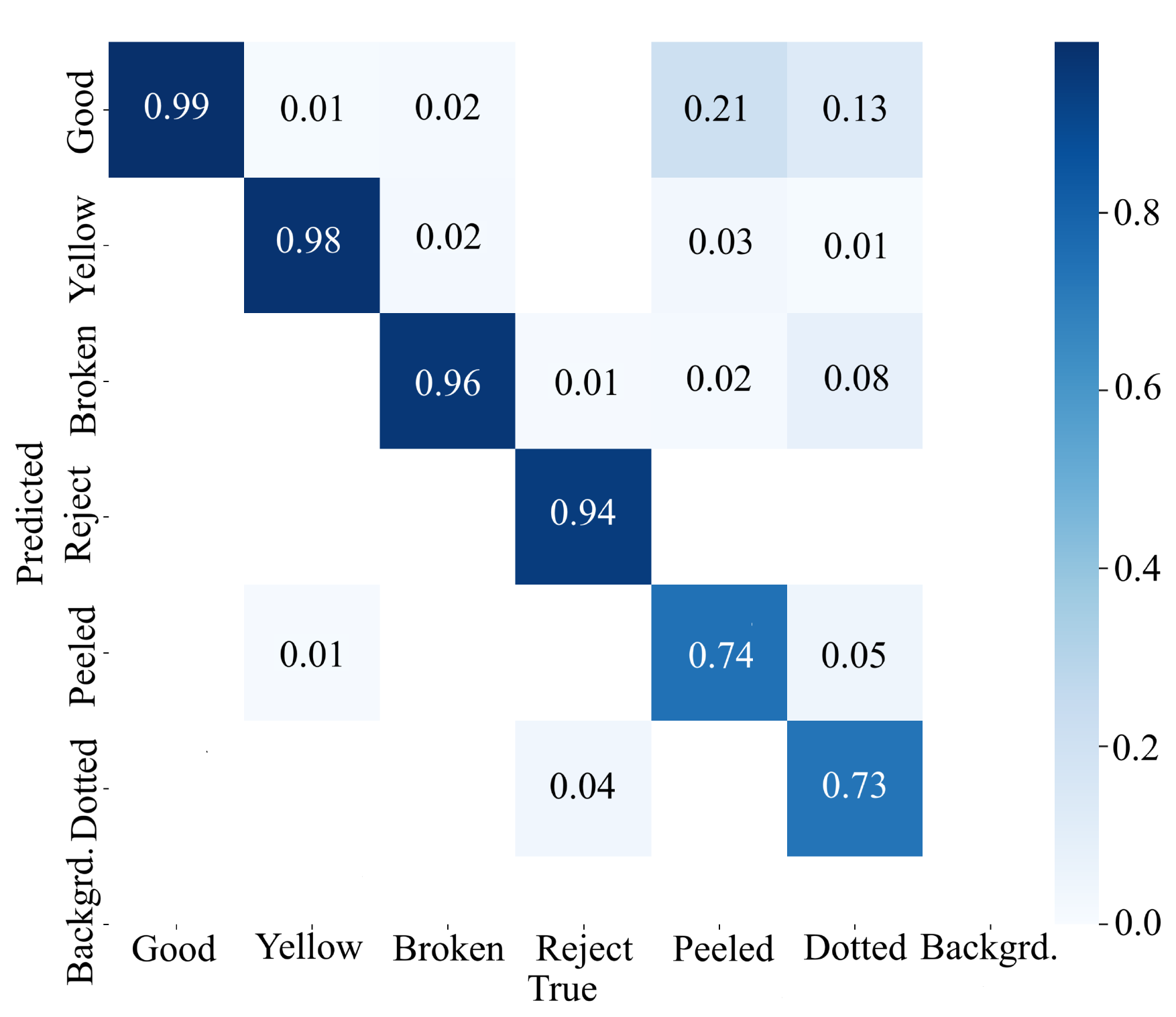}
  \caption{Confusion matrix obtained from the Yolo model for classification.}
  \label{fig:confusion-matrix}
\end{figure}

To further analyse the performance of the classification model beyond the general accuracy, we obtained the normalised confusion matrix shown in Figure \ref{fig:confusion-matrix}. This matrix reveals strong class-wise performance for classes Good to Reject, with true positive rates above 94\%. It also highlights certain weaknesses in distinguishing between Peeled and Dotted classes. These insights suggest that targeted improvements—such as refining data quality, increasing the number of samples for underperforming classes, or applying class-specific augmentation techniques—could further enhance the model’s robustness and overall classification reliability in more variable conditions.

\section{Conclusions}

The grain classification prototype operates effectively at a conveyor speed of 59 mm/s, achieving a separation accuracy of 87.2\% for moving lentils. The system follows a modular design that facilitates understanding and customisation, making it adaptable for performance optimisation and scaling in subsequent versions. Although its processing speed (8 grams per minute) is slower than that of commercial solutions, the prototype serves as a validation tool for future computer vision-based classification models. A demonstration video is included here \cite{authors2025}.

Optimising lighting by isolating the region of interest (ROI) and using fixed illumination significantly improved image capture. The YOLOv8 models for detection and classification achieved solid results, with an mAP50–95 of 0.8872 and an mAP50 of 0.9946 for detection, and a top-1 accuracy of 0.8920 and a top-5 accuracy of 1.00 for classification. The mechanical ejection system, featuring custom-designed nozzles and high-speed pneumatic solenoid valves, was reliably integrated with the vision system using an Arduino Nano board and the Firmata protocol. This prototype provides a foundation for future mechatronic systems that integrate conventional and emerging technologies—including AI—and are tailored to local needs.

\bibliographystyle{ieeetr}
\bibliography{bibliography}

\begin{thebibliography}{10}

\bibitem{fao2025}
{Food and Agriculture Organization of the United Nations}, ``{FAOSTAT},'' 2025.
\newblock Accessed: 2025-04-30.

\bibitem{indexbox2025}
W.~L.~M. Analysis, ``Index box,'' 2025.
\newblock Accessed: 2025-04-1.

\bibitem{shahin2003}
M.~Shahin and S.~Symons, ``Lentil type identification using machine vision,'' {\em Canadian Biosystems Engineering / Le génie des biosystèmes au Canada}, vol.~45, pp.~3.5--3.11, 2003.
\newblock Accessed: 2025-04-30.

\bibitem{wan2003}
J.~Wan and Y.~Lin, ``Rice quality classification using an automatic grain quality inspection system,'' {\em Transactions of the ASAE}, vol.~45, no.~2, pp.~146--158, 2003.
\newblock Accessed: 2025-04-30.

\bibitem{salcedo2024}
E.~Salcedo, C.~Huanca, and P.~Patzi, ``Enabling efficient royal quinoa quality inspection via mobile-based foreign body detection,'' in {\em 2024 IEEE Latin American Conference on Computational Intelligence (LA-CCI)}, pp.~1--6, 2024.

\bibitem{chen2019}
S.~Chen, J.~Xiong, W.~Guo, R.~Bu, Z.~Zheng, Y.~Chen, Z.~Yang, and R.~Lin, ``Colored rice quality inspection system using machine vision,'' {\em Journal of Cereal Science}, vol.~88, pp.~87--95, 2019.

\bibitem{delcoco2022}
M.~D. Coco, B.~Laddomada, G.~Romano, P.~Carcagnì, S.~K. Agrawal, and M.~Leo, ``Characterization of a collection of colored lentil genetic resources using a novel computer vision approach,'' {\em Foods}, vol.~11, no.~24, 2022.

\bibitem{pearson2010}
T.~C. Pearson, ``High-speed sorting of grains by color and surface texture,'' {\em Applied Engineering in Agriculture}, vol.~26, no.~3, pp.~499--505, 2010.

\bibitem{kundu2022}
N.~Kundu, G.~Rani, and V.~S. Dhaka, ``Seeds classification and quality testing using deep learning and yolo v5,'' in {\em Proceedings of the International Conference on Data Science, Machine Learning and Artificial Intelligence}, DSMLAI '21', (New York, NY, USA), p.~153–160, Association for Computing Machinery, 2022.

\bibitem{persak2020}
T.~Peršak, B.~Viltužnik, J.~Hernavs, and S.~Klančnik, ``Vision‑based sorting systems for transparent plastic granulate,'' {\em Applied Sciences}, vol.~10, no.~12, 2020.

\bibitem{terven2023}
J.~Terven, D.-M. Córdova-Esparza, and J.-A. Romero-González, ``A comprehensive review of yolo architectures in computer vision: From yolov1 to yolov8 and yolo-nas,'' {\em Machine Learning and Knowledge Extraction}, vol.~5, no.~4, pp.~1680--1716, 2023.

\bibitem{salcedo2022}
E.~Salcedo, M.~Jaber, and J.~Requena~Carrión, ``A novel road maintenance prioritisation system based on computer vision and crowdsourced reporting,'' {\em Journal of Sensor and Actuator Networks}, vol.~11, no.~1, 2022.

\bibitem{cvat2025}
C.~Corporation, ``Computer vision annotation tool (cvat),'' 2025.
\newblock Accessed: 2025-04-30.

\bibitem{authors2025}
D.~Rojas-Yana and E.~Salcedo, ``Demonstration video of the machine vision system.'' \url{https://youtu.be/ph6DICKLYtc}, July 2025.

\end{thebibliography}
\end{document}